\pdfoutput=1
\documentclass{article}

\PassOptionsToPackage{round,authoryear}{natbib}
\usepackage[preprint]{neurips_2026}
\usepackage[T1]{fontenc}
\usepackage{url}
\usepackage{booktabs}
\usepackage{amsmath,amsfonts,amssymb}
\usepackage{graphicx}
\usepackage{xcolor}
\usepackage{microtype}
\usepackage{multirow}
\usepackage{array}
\usepackage{enumitem}
\usepackage{placeins}

\newcolumntype{L}[1]{>{\raggedright\arraybackslash}p{#1}}
\newcounter{algorithm}
\renewcommand{\thealgorithm}{\arabic{algorithm}}
\makeatletter
\renewcommand{\@notice}{}
\makeatother

\title{\textsc{SMOPD}: Selective Token-Entropy Masking for Dirty-History Multi-Turn On-Policy Self-Distillation}

\author{
  \textbf{Chenyang Jiang}\thanks{Equal contribution.}
  \quad
  \textbf{Changhan Huang}\footnotemark[1]\\
  South China Normal University
}

\begin{document}
\maketitle
\thispagestyle{plain}

\begin{abstract}
Dirty-history rollouts make multi-turn on-policy self-distillation (OPSD) brittle: once a student emits an erroneous intermediate reply, later turns are conditioned on that reply, and uniform distillation can spend loss on tokens that carry little corrective signal. We introduce \textsc{SMOPD} (Selective Masking for On-Policy Distillation), a loss-only stabilization method for multi-turn OPSD. For each generated middle-turn reply, \textsc{SMOPD} ranks token positions by student entropy and removes the lowest-entropy 20\% from the clipped generalized Jensen-Shannon distillation loss; final-answer and FULL-preservation losses are unchanged. This design targets token-level uncertainty rather than coarse trajectory outcomes, adds no parameters, and has zero inference-time overhead. We compare \textsc{SMOPD} with a correctness-scaling variant that multiplies a common detached reliability proxy using final-answer correctness. On LiC with Qwen3 models, \textsc{SMOPD} improves SHARDED-view accuracy by 1.0--2.5 percentage points in single-seed 1.7B, 4B, and 8B comparisons, and a small 4B multi-seed check shows a +1.7pp mean SHARDED gain over baseline (two-tailed $p=0.022$). Adding the outcome scalar is harmful without masking at 1.7B ($-$4.0pp) and remains scale-dependent when combined with masking (+1.3pp at 4B, neutral at 1.7B, and $-$0.5pp at 8B). These archived aggregate results suggest that token-level uncertainty is a more reliable stabilization signal than scalar final-answer correctness in this evaluated dirty-history OPSD setting, while leaving causal mechanism tests and broader benchmark validation to future work.
\end{abstract}

\section{Introduction}

Large language models (LLMs) are increasingly deployed in multi-turn interactive settings---chat, tool use, and collaborative problem solving---where they must maintain coherence across multiple exchanges. Training LLMs for such settings is challenging: supervised fine-tuning requires costly human demonstrations at every turn, while single-turn evaluation is a poor proxy for multi-turn performance~\citep{dubois2024length,madaan2024self}.

On-Policy Self-Distillation (OPSD) has emerged as an effective paradigm for multi-turn training. The model generates its own trajectories (on-policy rollouts), and a frozen EMA teacher provides distillation targets under clean, single-turn context. MAIGO~\citep{zheng2025maigo} implements this via a history-cleaning mechanism that isolates the teacher from student-generated errors, plus a single-turn preservation branch to prevent catastrophic forgetting.

However, a fundamental challenge remains: \textbf{dirty history}. During multi-turn rollouts, the student accumulates errors across turns. Each intermediate response conditions the next prompt, creating a compounding distortion that pulls the student away from the reference distribution. Prior work has proposed various techniques for single-turn settings---RL stabilization (PPO clipping~\citep{schulman2017proximal}), preference optimization (DPO~\citep{rafailov2024direct}, KTO~\citep{ethayarajh2024kto}), and distillation-specific stabilization (REOPOLD~\citep{ko2026reopold}, EOPD~\citep{jin2025eopd})---but whether any of these transfer to multi-turn OPSD is an open question.

Against this backdrop, we define a specific, reusable stabilization component for dirty-history OPSD: percentile-based student-entropy selection inside the middle-turn loss. We call this entropy-masked variant \textsc{SMOPD} (Selective Masking for On-Policy Distillation). The selection rule is closest to REOPOLD's entropy-guided dynamic sampling~\citep{ko2026reopold}; our contribution is to place it in a multi-turn clean-teacher / dirty-student GJS distillation loss and compare it directly with coarse trajectory-outcome weighting.

We study two interventions:

\begin{enumerate}
    \item \textbf{Token-level entropy masking}: Mask low-entropy (overconfident) tokens from the GJS distillation loss. The intuition: when the student assigns high confidence to a token, that token contributes little learning signal and may represent overconfident errors formed under dirty history.

    \item \textbf{Outcome-guided correctness scaling}: Multiply the common reliability proxy by a scalar that up-weights middle turns when the final answer is incorrect and down-weights them when correct. The intuition: failed trajectories may carry more informative signal about where dirty-history corruption occurs; successful trajectories have less to learn from.
\end{enumerate}

We evaluate both via systematic ablation across three model scales (Qwen3-1.7B, 4B, 8B) on the LiC multi-turn benchmark. Our findings are:

\begin{itemize}
    \item \textbf{Entropy masking is the most consistent positive signal in our evidence}: It improves SHARDED-view accuracy by 1.0--2.5pp across the evaluated model scales, with zero additional parameters and zero inference overhead.

    \item \textbf{Outcome correctness scaling degrades without masking and is mixed when combined}: Without masking, it lowers SHARDED by 4.0pp. Combined with masking, results are scale-dependent: neutral at 1.7B, +1.3pp at 4B (two-tailed $p=0.101$; directional one-tailed $p=0.0504$), and $-$0.5pp at 8B. The combination does not reliably outperform masking alone.

    \item \textbf{Reliability evidence is strongest for entropy masking at 4B}: The relative gain from entropy masking remains positive at 200 steps, and the 4B multi-seed comparison shows a +1.7pp SHARDED mean gain with low observed seed-to-seed spread.
\end{itemize}

Our results suggest a practical starting point for multi-turn OPSD: prioritize token-level uncertainty (entropy-based token selection) before adding coarse outcome signals. The simplicity of entropy masking makes it straightforward to test in OPSD pipelines.

In summary, our contributions are:
\begin{itemize}
    \item We introduce \textsc{SMOPD}, a loss-only selective masking component that adapts percentile-based token-level entropy selection to multi-turn OPSD with dirty-history rollouts and a GJS middle-turn loss.
    \item We analyze token-level confidence as a motivation for loss masking and separate this mechanism hypothesis from the archived aggregate accuracy evidence.
    \item Through ablation across three model scales, we find that entropy masking improves multi-turn accuracy by 1.0--2.5pp in our evaluated settings, while correctness-based outcome scaling---effective in single-turn RL-style settings---does not transfer reliably in these experiments.
    \item We provide a small multi-seed check at 4B that supports the entropy-masking signal and indicates that the outcome scalar's effect remains uncertain across scales.
    \item We identify a methodological concern---spurious large effects from single-seed OPSD experiments---and offer practical evaluation recommendations.
\end{itemize}

\section{Related Work}

\begin{table}[t]
\centering
\small
\caption{Positioning relative to closely related OPD/OPSD stabilization methods.}
\label{tab:related_positioning}
\begin{tabular}{@{}L{0.16\textwidth}L{0.23\textwidth}L{0.17\textwidth}L{0.32\textwidth}@{}}
\toprule
\textbf{Method} & \textbf{Primary signal} & \textbf{Setting} & \textbf{Relation to this paper} \\
\midrule
MAIGO~\citep{zheng2025maigo} & Clean-history teacher references & Multi-turn LiC OPSD & Provides the training family and dirty-history motivation. \\
REOPOLD~\citep{ko2026reopold} & Student-entropy token selection & Single-turn OPD refinement & Closest token-selection predecessor; we test the idea in middle-turn GJS OPSD. \\
EOPD~\citep{jin2025eopd} & Teacher-entropy KL switching & Single-turn OPD & Modifies divergence choice rather than masking low-entropy student tokens. \\
SCOPE~\citep{zheng2026scope} & Correctness-routed adaptive weighting & OPD reasoning alignment & Uses richer dual-path weighting; our scalar outcome weighting is a simpler stress test. \\
AOPD~\citep{jia2026aopd} & Advantage-region token intervention & Token-level OPD & Switches learning mode by advantage; our mask uses per-reply student-entropy percentiles. \\
UniSD~\citep{unisd2026} & Agreement, EMA, clipping, auxiliary losses & General self-distillation & Broader reliability-aware framework; our study isolates one lightweight loss mask. \\
CREDIT~\citep{shen2026credit} & Input-specific credit signal & OPSD / OPD credit assignment & Supports the view that coarse trajectory outcomes are insufficiently granular. \\
OGLS-SD~\citep{yang2026ogls} & Outcome-guided logit steering & OPSD reasoning & Closest outcome-guided counterpart; our outcome variant is a weaker scalar baseline. \\
\bottomrule
\end{tabular}
\end{table}

Table~\ref{tab:related_positioning} is the main positioning device for this paper. The key distinction is scope: \textsc{SMOPD} is a narrow loss-side intervention inside a MAIGO-style clean-teacher / dirty-student training family, not a replacement for broader OPD frameworks or richer outcome-guided credit-assignment methods.

\paragraph{Multi-turn OPSD.}
OPSD trains on student-generated trajectories while an EMA teacher supplies reference logits~\citep{zhao2026opsd}. MAIGO~\citep{zheng2025maigo} extends this idea to multi-turn LiC by cleaning teacher history and preserving FULL-view competence. We use the same clean-teacher / dirty-student motivation, but average all eligible middle-turn losses and study selective masking inside the GJS loss. Adjacent self-improvement and self-distillation lines include self-rewarding and self-play training~\citep{yuan2024self,chen2024self}, SDFT~\citep{shenfeld2026sdft}, context-conditioned OPD~\citep{ye2026onpolicy}, empirical OPD failure analyses~\citep{fu2025revisiting}, reflective OPSD~\citep{zhao2026rosd}, and reliability-aware self-distillation frameworks~\citep{song2026survey,unisd2026}.

\paragraph{Stabilization signals.}
Single-turn RLHF and preference optimization rely on KL-style constraints or preference-derived rewards~\citep{ouyang2022training,rafailov2024direct,ethayarajh2024kto}; G-OPD and ExOPD connect OPD to dense KL-constrained RL~\citep{yang2026gopd}. Token-level OPD methods are closer to our mechanism: REOPOLD selects high-entropy student tokens during refinement~\citep{ko2026reopold}, EOPD switches KL direction using entropy~\citep{jin2025eopd}, AOPD changes token-level learning mode by advantage region~\citep{jia2026aopd}, and DRKL studies reverse-KL overconfidence~\citep{luong2026drkl}. We adapt percentile-based student-entropy selection to dirty-history middle turns, with the percentile computed within each generated reply rather than over a batch.

\paragraph{Outcome and benchmark context.}
Process rewards provide step-level feedback but require extra supervision~\citep{lightman2023let}. Outcome-guided methods such as OGLS-SD and SCOPE use richer steering or dual-path weighting~\citep{yang2026ogls,zheng2026scope}; CREDIT argues that coarse trajectory outcomes are insufficient for OPD credit assignment~\citep{shen2026credit}. Our outcome variant is intentionally weaker: a scalar correctness-dependent multiplier on a common reliability-weighted middle-turn loss. We evaluate on LiC~\citep{laban2025lic}, whose paired FULL/SHARDED protocol directly measures the multi-turn degradation that MAIGO targets. MT-Bench offers complementary multi-turn evaluation~\citep{zheng2024judging}, but lacks this paired-view control.

\paragraph{Entropy signal.}
Entropy regularization is standard in reinforcement learning~\citep{haarnoja2018soft,schulman2017proximal}, and recent LLM work links entropy dynamics to reasoning diversity and exploration~\citep{cui2025entropy,cheng2025reasoning}. In contrast, \textsc{SMOPD} uses entropy as a per-token selection signal for middle-turn distillation, not as a global regularizer.

\section{Motivation: Token-Level Confidence in Multi-Turn OPSD}

Dirty-history OPSD averages a per-token GJS loss over student-generated middle turns, but the tokens in a reply are not equally informative. Low-entropy positions often correspond to predictable continuations, formatting, or locally committed choices; high-entropy positions more often mark decision points such as operators, variable bindings, or logical connectors. Under dirty history, confidence can also become miscalibrated after earlier student errors enter the context.

\textsc{SMOPD} therefore treats entropy masking as a stabilization hypothesis: retain the uncertain token positions where correction is more likely to matter, and remove the lowest-entropy positions from the middle-turn loss. We use a per-reply percentile rather than a global threshold because entropy scales shift across tasks, turns, and training stages. Throughout the reported experiments, $\beta=0.8$ retains the highest-entropy 80\% of tokens within each generated middle-turn reply. The aggregate results below test whether this rule improves accuracy; they do not by themselves prove a token-level causal mechanism.

\section{Method}

\begin{figure}[t]
\centering
\includegraphics[width=0.95\textwidth]{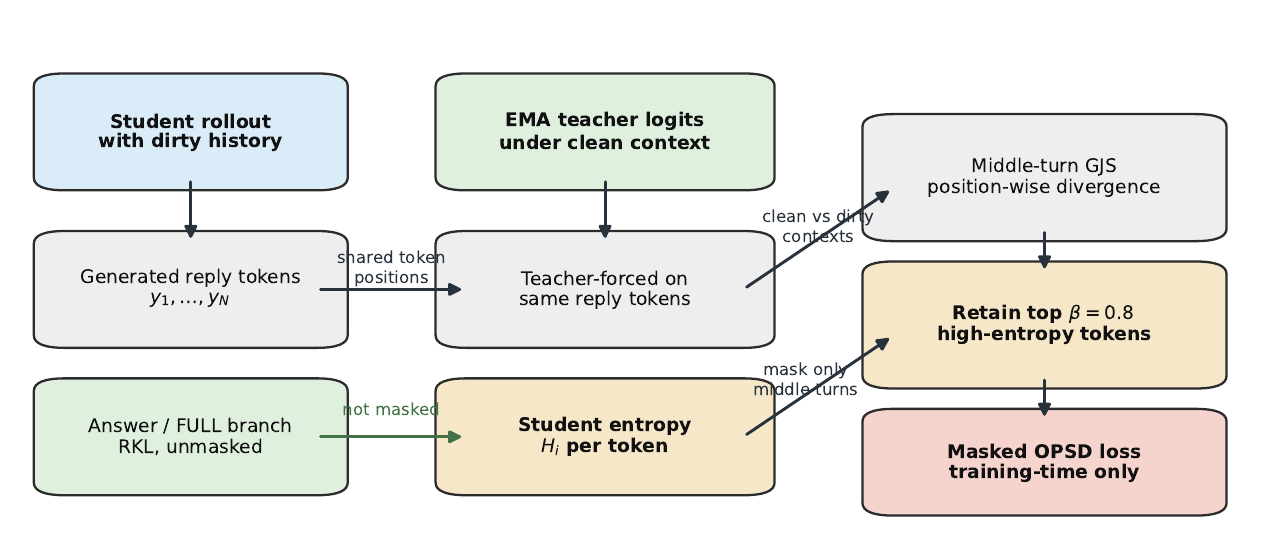}
\caption{\textbf{Entropy masking in multi-turn OPSD.} The student generates dirty-history rollouts; the EMA teacher is evaluated under clean context on the same generated response tokens. Entropy masking is applied only to middle-turn GJS losses, while answer-turn and FULL-preservation RKL losses remain unmasked.}
\label{fig:method_schematic}
\end{figure}

We adopt a MAIGO-inspired clean-teacher / dirty-student framework~\citep{zheng2025maigo}: the student model $\pi_\theta$ generates multi-turn rollouts with dirty history, while a frozen EMA teacher $\pi_{\text{EMA}}$ produces token-level logits under clean, single-turn context---middle turns see only user turns $u_1, \ldots, u_t$, and the answer turn sees the canonical FULL-view specification $f$. Our implementation deliberately differs from MAIGO in two places: it averages over all eligible middle turns in a rollout rather than sampling one eligible turn as a single-turn estimator, and it uses the detached reliability proxy in \S\ref{sec:method_outcome} as the common middle-turn weight. Outcome-guided variants multiply that proxy by a final-answer correctness scalar. Figure~\ref{fig:method_schematic} summarizes where entropy masking enters the training loop.

\subsection{Preliminaries}

Given a student-generated reply at turn $t$ with tokens $y_1, \ldots, y_N$, the per-token loss at position $i$ is computed as a divergence between the student's distribution $p_{\text{student}}^i = \pi_\theta(\cdot \mid c_i^{\text{dirty}})$ and the teacher's distribution $p_{\text{teacher}}^i = \pi_{\text{EMA}}(\cdot \mid c_i^{\text{clean}})$, where $c_i^{\text{dirty}}$ and $c_i^{\text{clean}}$ denote the dirty-history and clean-history contexts, respectively.

\textbf{Token alignment and teacher forcing}. For each generated student reply, both student and teacher distributions are evaluated on the same token sequence $y_1,\ldots,y_N$. The contexts differ, but the supervised positions do not: the student distribution is conditioned on the dirty prefix that includes prior student replies, while the teacher distribution is teacher-forced on the student-generated continuation under the corresponding clean context. This avoids variable-length alignment between independently generated teacher and student replies; divergence is computed position-wise over the shared vocabulary at each generated token position.

\textbf{Generalized Jensen-Shannon Divergence}. For middle turns, we use GJS with coefficient $\beta_{\text{mid}} = 0.5$:
\begin{equation}
\label{eq:gjs}
    \text{GJS}(p_{\text{student}}^i, p_{\text{teacher}}^i) = \beta_{\text{mid}}\, D_{\text{KL}}(p_{\text{teacher}}^i \,\|\, m) + (1-\beta_{\text{mid}})\, D_{\text{KL}}(p_{\text{student}}^i \,\|\, m)
\end{equation}
where $m = \beta_{\text{mid}} \, p_{\text{teacher}}^i + (1-\beta_{\text{mid}}) \, p_{\text{student}}^i$ is the interpolated mixture distribution. GJS provides symmetric, bounded gradients suitable for intermediate reasoning steps, avoiding the mode-seeking bias of pure reverse KL.

\textbf{Reverse KL Divergence}. For the answer turn, we use RKL with coefficient $\beta_{\text{final}} = 1.0$:
\begin{equation}
\label{eq:answer_rkl}
    \mathcal{L}_{\text{answer}} = \frac{1}{N}\sum_{i=1}^{N} \widetilde{D}_{\text{KL}}(p_{\text{student}}^i \,\|\, p_{\text{teacher}}^i)
\end{equation}
RKL penalizes student samples outside the teacher's support---appropriate for the final answer where precision matters.

\textbf{Pointwise clipping, preservation, and EMA}. Following MAIGO's token-level clipping setting, every per-token GJS and RKL term used for training is clipped at $\epsilon_{\text{clip}}=0.5$:
\begin{equation}
\label{eq:pointwise_clipping}
    \widetilde{d}_i = \min(d_i, \epsilon_{\text{clip}}).
\end{equation}
We write $\widetilde{\text{GJS}}$ and $\widetilde{D}_{\text{KL}}$ for clipped divergences. With probability $\rho = 0.2$, we also compute FULL-branch RKL on the canonical single-turn task $f$. After each optimizer step, the EMA model is updated as $\theta_{\text{EMA}} \leftarrow \lambda \theta_{\text{EMA}} + (1-\lambda) \theta$ with $\lambda = 0.99$.

\subsection{Token-Level Entropy Masking}

\label{sec:method_entropy}

For each middle turn, the GJS divergence produces a per-token loss $\ell_i$ on the student's generated reply. We compute student entropy at each token position:

\begin{equation}
\label{eq:token_entropy}
    H_i = -\sum_{v \in V} p_\theta(v | x_{<i}) \log p_\theta(v | x_{<i})
\end{equation}

where $V$ is the vocabulary. Given a retention ratio $\beta \in (0, 1]$, we retain the top $\beta$ fraction of tokens by entropy and zero out the lowest-entropy positions:

\begin{equation}
\label{eq:masked_mid_loss}
    \mathcal{L}_{\text{mid},t} = \frac{1}{|\mathcal{T}_t|} \sum_{i \in \mathcal{T}_t} \widetilde{\text{GJS}}(p_{\text{student}}^i, p_{\text{teacher}}^i)
\end{equation}

where $\mathcal{T}_t = \{i : H_i \geq \text{quantile}(\{H_j\}_{j=1}^{N_t}, 1-\beta)\}$ is the retained token set within reply $t$. We use $\beta = 0.8$. The operation is loss-side only: no extra parameters, no generation changes, and zero inference-time overhead.

\subsection{Outcome-Guided Turn Weighting}

\label{sec:method_outcome}

For each middle turn $t$, we compute a detached reliability proxy. This reliability weighting is shared by all reported configurations; when outcome guidance is enabled, we additionally apply a correctness-dependent scalar. Let $y_i$ be the $i$th token in the student-generated middle reply. We measure the student--teacher discrepancy on the realized tokens as

\begin{equation}
\label{eq:reliability_delta_proxy}
    \delta_t^{\text{proxy}} = \frac{1}{N_t}\sum_{i=1}^{N_t} \left| \log p_{\text{student}}^i(y_i) - \log p_{\text{teacher}}^i(y_i) \right|.
\end{equation}

Let $\tilde{\delta}$ be the median of $\max(\delta_j^{\text{proxy}}, \epsilon)$ over eligible middle turns in the same rollout. We define a MAIGO-inspired reliability proxy

\begin{equation}
\label{eq:reliability_proxy}
    r_t = \frac{\tilde{\delta}}{\tilde{\delta} + \max(\delta_t^{\text{proxy}}, \epsilon)}.
\end{equation}

This proxy follows MAIGO's motivation~\citep{zheng2025maigo} of down-weighting turns that have drifted far from the clean-context reference, but it is not an exact reproduction of MAIGO's adaptive estimator. We then multiply it by a correctness-dependent scalar:

\begin{equation}
\label{eq:outcome_weight}
    w_t = \operatorname{clip}_{[0,1]}\left(r_t \cdot \bigl(1 + \eta \cdot (1 - 2o)\bigr)\right),
\end{equation}

where $o \in \{0, 1\}$ is the binary outcome (1 = correct, 0 = incorrect). We use $\eta = 0.2$, so incorrect outcomes multiply the reliability proxy by 1.2 and correct outcomes by 0.8.

The combined middle-turn loss with both techniques becomes:

\begin{equation}
\label{eq:combined_mid_loss}
    \mathcal{L}_{\text{mid}}^{\text{combined}} =
    \frac{1}{|\mathcal{I}|} \sum_{t \in \mathcal{I}} w_t
    \left(
    \frac{1}{|\mathcal{T}_t|} \sum_{i \in \mathcal{T}_t}
    \widetilde{\text{GJS}}(p_{\text{student}}^i, p_{\text{teacher}}^i)
    \right)
\end{equation}

where $\mathcal{I}$ is the set of eligible non-empty middle turns. MAIGO's Algorithm 1 samples one eligible middle turn as an estimator; our implementation averages all eligible middle-turn losses in the rollout.

\subsection{Training Procedure}

\label{sec:algorithm}

Algorithm~\ref{alg:training} summarizes the training loop. Following the REOPOLD-inspired schedule, entropy masking is disabled during the first 33\% of steps by setting $\beta = 1.0$; unlike REOPOLD, we do not use a reward-filtering phase.

\begin{figure}[tbp]
\centering
\refstepcounter{algorithm}
\label{alg:training}
\begin{minipage}{0.96\textwidth}
\small
\hrule\vspace{3pt}
\textbf{Algorithm \thealgorithm: Entropy-masked multi-turn OPSD training.}\\
\textbf{Input:} Student $\pi_\theta$, EMA teacher $\pi_{\text{EMA}}$, training set $\mathcal{D}$, masking ratio $\beta = 0.8$, outcome sensitivity $\eta$, exploration fraction $\gamma = 0.33$.\\
\textbf{Output:} Trained student $\pi_\theta$.
\vspace{2pt}
\begin{enumerate}[leftmargin=1.6em,itemsep=1pt,topsep=2pt]
\item Initialize $\pi_{\text{EMA}} \leftarrow \pi_\theta$.
\item For step $s = 1,\ldots,S$, sample $\mathcal{B}\sim\mathcal{D}$ and generate dirty-history rollouts $\{a_1,\ldots,a_T\}\sim\pi_\theta$.
\item Compute teacher logits $\pi_{\text{EMA}}(\cdot\mid\text{clean context})$ for each turn.
\item If $s>\gamma S$, compute $H_i$ (Eq.~\ref{eq:token_entropy}) and retain the per-reply top-$\beta$ entropy tokens; otherwise set $\mathcal{T}_t$ to all tokens.
\item Compute reliability proxies $r_t$ and set $w_t$ by Eq.~\ref{eq:outcome_weight}, with $w_t=r_t$ when $\eta=0$.
\item Clip token losses (Eq.~\ref{eq:pointwise_clipping}) and compute $\mathcal{L}_{\text{mid}}^{\text{combined}}$ (Eq.~\ref{eq:combined_mid_loss}) on retained tokens.
\item Add answer-turn RKL and, with probability $\rho=0.2$, FULL-branch RKL.
\item Update $\pi_\theta$ by gradient descent and update EMA using decay $\lambda$.
\end{enumerate}
\vspace{3pt}\hrule
\end{minipage}
\end{figure}

\section{Experimental Setup}

\subsection{Models and Optimization}

We use Qwen3-Instruct models~\citep{qwen2025qwen3} at three scales: 1.7B, 4B, and 8B parameters. All models are fine-tuned with LoRA~\citep{hu2022lora} (rank $r=64$, $\alpha=128$, dropout $0.0$) applied to all linear projections in the attention and feed-forward layers.

We train for 100 steps using AdamW~\citep{loshchilov2019decoupled} with learning rate $5\times 10^{-6}$. EMA decay is set to $\lambda = 0.99$. The FULL preservation branch is sampled with probability $\rho = 0.2$. Following the REOPOLD-inspired schedule described in \S\ref{sec:algorithm}, an exploration phase (entropy masking disabled, without REOPOLD's reward-based filtering) covers the first 33\% of training steps. All experiments use seed 42 unless noted otherwise.

\subsection{Benchmark and Metric}

We use the LiC (Language in Context) combined benchmark. The training set contains 12,165 multi-turn examples spanning math, database, code, and action-planning domains. Evaluation uses 200 held-out math examples with both FULL-view (single-turn complete task) and SHARDED-view (step-by-step turn revelation) protocols. The primary metric is \emph{SHARDED accuracy}, which measures multi-turn reasoning performance under dirty-history conditions.

\subsection{Evidence Boundary and Provenance}

All tabulated and plotted accuracy values are aggregate metrics traceable to archived run artifacts. We exclude unarchived token-level diagnostics, per-turn slices, prompt-level examples, and exploratory parsing outputs from the quantitative evidence base. Accordingly, $\beta=0.8$ is treated as a fixed operating point rather than a proven optimum. The arXiv source package contains only manuscript source, style file, and figures; it excludes model weights, checkpoints, logs, and raw evaluation outputs.

\subsection{Configurations}

We test four configurations:
\begin{itemize}
    \item \textbf{baseline}: Clean-teacher / dirty-student OPSD with the common reliability proxy, no entropy mask and no outcome scalar ($\beta = 1.0$, $\eta = 0.0$).
    \item \textbf{+entropy}: Entropy masking on top of the common reliability proxy ($\beta = 0.8$, $\eta = 0.0$).
    \item \textbf{+outcome}: Correctness scalar on top of the common reliability proxy ($\beta = 1.0$, $\eta = 0.2$).
    \item \textbf{combined}: Both stabilization techniques active ($\beta = 0.8$, $\eta = 0.2$). Two archived training runs with identical hyperparameters are averaged to estimate run-to-run variation.
\end{itemize}

\subsection{Hardware}

All experiments run on a single NVIDIA RTX PRO 6000 Blackwell GPU (96GB VRAM). Each 100-step training run takes approximately 40--50 minutes; full evaluation (200 samples $\times$ 2 views) takes an additional 60--80 minutes.

\section{Results}

\subsection{Main Ablation Study (1.7B)}

\label{sec:results_ablation}

\begin{table}[t]
\centering
\caption{Ablation study on Qwen3-1.7B (100 steps, seed=42).}
\label{tab:ablation}
\begin{tabular}{lccccc}
\toprule
\textbf{Config} & $\beta$ & $\eta$ & \textbf{FULL} & \textbf{SHARDED} & $\Delta$SH \\
\midrule
baseline & 1.0 & 0.0 & 76.5\% & 66.5\% & -- \\
+entropy & 0.8 & 0.0 & 79.0\% & \textbf{69.0\%} & +2.5 \\
+outcome & 1.0 & 0.2 & 78.5\% & 62.5\% & $-$4.0 \\
combined$^{\,\dagger}$ & 0.8 & 0.2 & 76.5\% & 65.8\% & $-$0.7 \\
\bottomrule
\end{tabular}

\vspace{2pt}
{\footnotesize $^\dagger$Mean of two archived training runs with identical hyperparameters (75.0\%/64.5\% and 78.0\%/67.0\%). The 2.5pp SHARDED range illustrates the scale of run-to-run variation in this setting.}
\end{table}

\begin{figure}[tbp]
\centering
\includegraphics[width=0.85\textwidth]{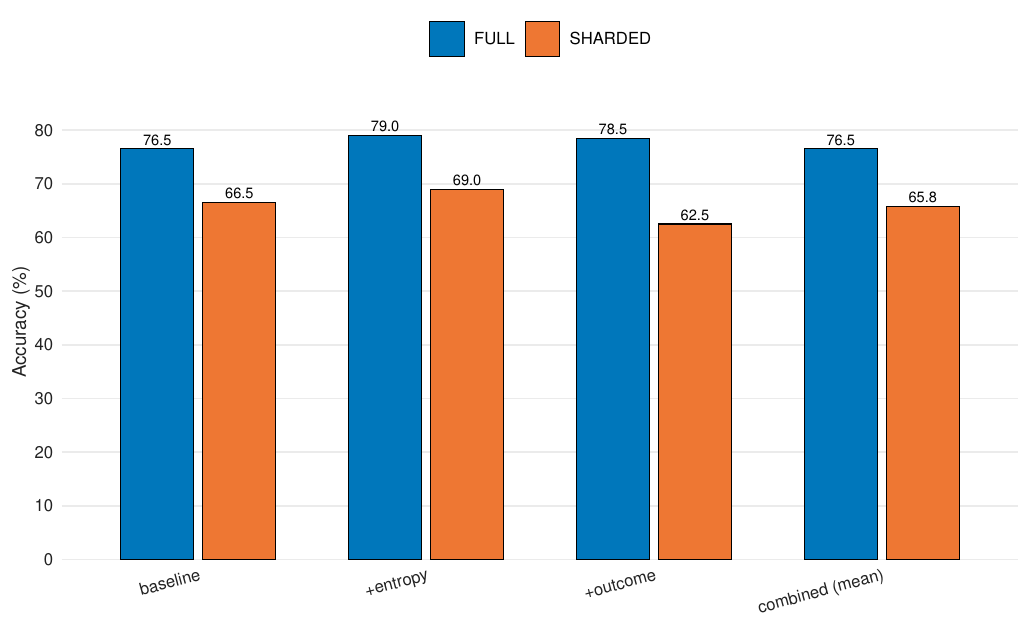}
\caption{\textbf{Ablation results on Qwen3-1.7B.} In this ablation, entropy masking raises SHARDED by 2.5pp, the outcome scalar lowers it by 4.0pp, and the combined mean does not improve over baseline.}
\label{fig:ablation}
\end{figure}

Table~\ref{tab:ablation} presents the core ablation results. Entropy masking gives the largest positive change in this ablation: it improves SHARDED accuracy by 2.5pp over baseline alongside a comparable FULL improvement (+2.5pp). Adding the outcome scalar without masking lowers SHARDED by 4.0pp, suggesting that binary outcome signals are too noisy to guide middle-turn training in this setting.

The combined configuration ($\beta=0.8, \eta=0.2$) was run twice with identical hyperparameters. The two runs differ by 2.5pp SHARDED (64.5\% vs.\ 67.0\%), illustrating the scale of run-to-run variation in this setting. The mean result (65.8\%) is 0.7pp below baseline, suggesting that combining the outcome scalar with entropy masking provides no clear benefit over masking alone at the 1.7B scale and may slightly degrade performance.

\subsection{Cross-Scale Validation (4B, 8B)}

\label{sec:results_cross_scale}

We probe whether the 1.7B findings transfer to larger models by running baseline, +entropy, and combined on Qwen3-4B and Qwen3-8B.

\begin{table}[t]
\centering
\caption{Cross-scale validation of entropy masking (100 steps, seed=42). $\Delta$SH = SHARDED gain over baseline at each scale.}
\label{tab:cross_scale}
\begin{tabular}{llccc}
\toprule
\textbf{Model} & \textbf{Config} & \textbf{FULL} & \textbf{SHARDED} & $\Delta$SH \\
\midrule
\multirow{2}{*}{1.7B} & baseline & 76.5\% & 66.5\% & -- \\
                       & +entropy & 79.0\% & 69.0\% & +2.5 \\
\midrule
\multirow{2}{*}{4B}   & baseline & 78.5\% & 64.5\% & -- \\
                       & +entropy & 79.5\% & 66.0\% & +1.5 \\
\midrule
\multirow{2}{*}{8B}   & baseline & 81.0\% & 74.0\% & -- \\
                       & +entropy & 81.5\% & 75.0\% & +1.0 \\
\bottomrule
\end{tabular}
\end{table}

\begin{figure}[tbp]
\centering
\includegraphics[width=0.80\textwidth]{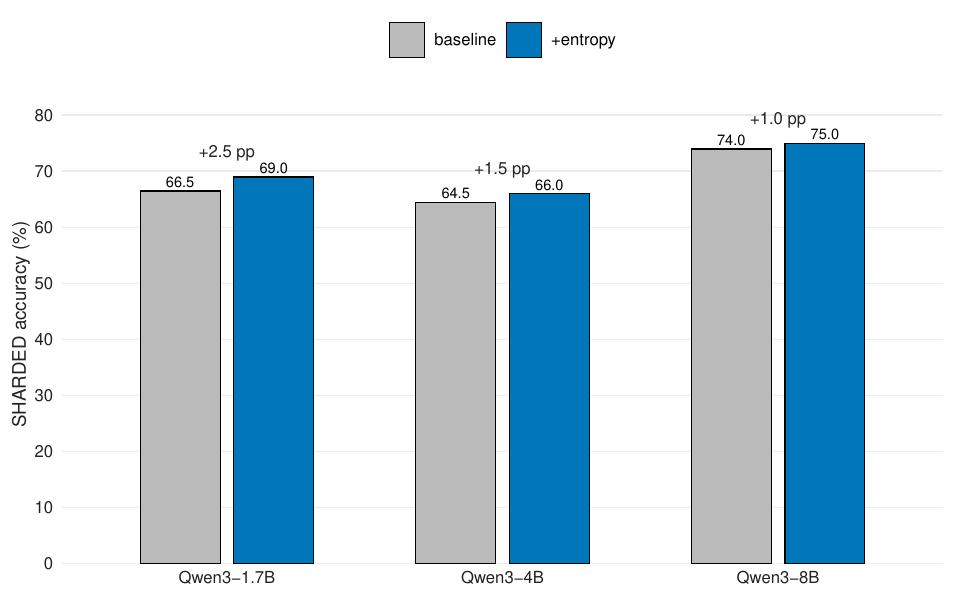}
\caption{\textbf{Single-seed cross-scale comparisons.} Entropy masking has positive SHARDED deltas at all evaluated scales, with smaller observed gains at larger model sizes.}
\label{fig:cross_scale}
\end{figure}

Table~\ref{tab:cross_scale} shows positive single-seed SHARDED deltas for entropy masking at all evaluated scales: +2.5pp (1.7B), +1.5pp (4B), and +1.0pp (8B). The gains are small and should be read descriptively, but their direction is consistent with the 1.7B ablation. Combined results remain scale-dependent: $-$0.7pp at 1.7B, +1.3pp at 4B, and $-$0.5pp at 8B.

\begin{table}[t]
\centering
\caption{Absolute SHARDED accuracy for combined-configuration comparisons. The 1.7B combined value is the mean of two archived runs, the 4B row uses the multi-seed means from Table~\ref{tab:multiseed}, and the 8B row is a single-seed comparison.}
\label{tab:combined_scale}
\begin{tabular}{lcccc}
\toprule
\textbf{Model} & \textbf{Baseline} & \textbf{+entropy} & \textbf{Combined} & $\Delta$SH vs baseline \\
\midrule
1.7B & 66.5\% & 69.0\% & 65.8\% & $-$0.7 \\
4B   & 64.2\% & 65.8\% & 65.5\% & +1.3 \\
8B   & 74.0\% & 75.0\% & 73.5\% & $-$0.5 \\
\bottomrule
\end{tabular}
\end{table}

\textbf{Addressing a 4B outlier}. An initial 4B combined run produced a +7.0pp SHARDED gain (71.5\% vs.\ 64.5\%), but replication reduced the estimate to +1.3pp mean gain. We treat the outlier as evidence of run/seed variance rather than as the main effect.

\begin{figure}[tbp]
\centering
\includegraphics[width=0.78\textwidth]{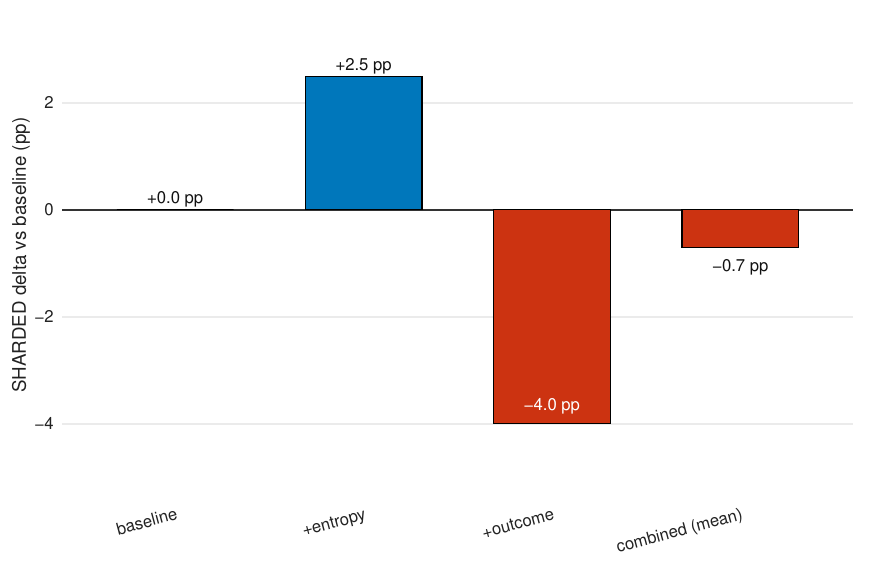}
\caption{\textbf{Effect breakdown (1.7B).} Entropy masking gives the largest positive SHARDED delta in this ablation; the outcome scalar without masking is negative; combined does not improve over baseline at this scale.}
\label{fig:delta}
\end{figure}

\subsection{Small Multi-Seed Reliability Check}

\label{sec:results_reliability}

For 4B baseline and +entropy, we combine the cross-scale seed-42 runs with seeds 123 and 456. For combined, the initial seed-42 outlier is reported separately; Table~\ref{tab:multiseed} uses three subsequent reliability runs.

\begin{table}[t]
\centering
\caption{Multi-seed validation on 4B, 100 steps. Mean $\pm$ std across 3 seeds. Table $p$-values are from Welch's two-tailed $t$-test comparing SHARDED against baseline; directional one-tailed values are reported in text.}
\label{tab:multiseed}
\begin{tabular}{lcccc}
\toprule
\textbf{Config} & \textbf{FULL (mean$\pm$std)} & \textbf{SHARDED (mean$\pm$std)} & $\Delta$SH & $p_{\text{2-tail}}$ \\
\midrule
baseline & 78.2$\pm$0.6\% & 64.2$\pm$0.6\% & -- & -- \\
+entropy & 79.7$\pm$0.3\% & \textbf{65.8}$\pm$\textbf{0.3\%} & +1.7 & 0.022 \\
combined$^{\,\ddagger}$ & 78.5$\pm$0.9\% & 65.5$\pm$0.9\% & +1.3 & 0.101 \\
\bottomrule
\end{tabular}
\vspace{2pt}
{\footnotesize $^\ddagger$The combined row uses three subsequent reliability runs. The earlier combined seed-42 outlier is reported separately in \S\ref{sec:outlier}.}
\end{table}

Entropy masking shows a +1.7pp SHARDED mean gain (one-tailed $p=0.011$; two-tailed $p=0.022$). Combined is weaker (+1.3pp, one-tailed $p=0.0504$; two-tailed $p=0.101$) and inconsistent across scales. With only three seeds, these $p$-values are rough reliability checks rather than precise population estimates.

\begin{figure}[tbp]
\centering
\includegraphics[width=0.72\textwidth]{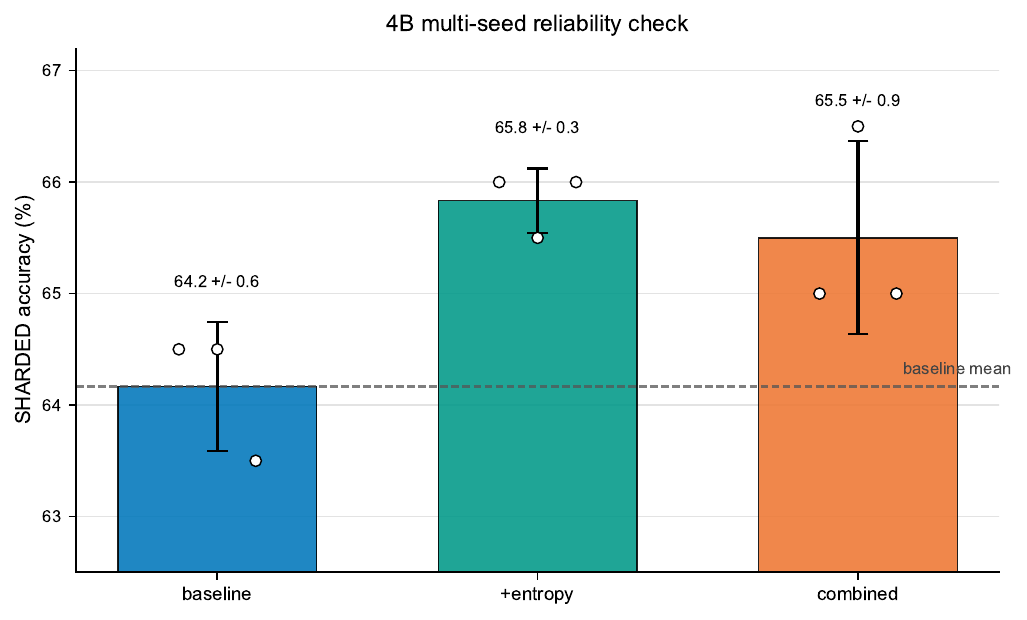}
\caption{\textbf{4B multi-seed reliability check.} Bars show mean SHARDED accuracy across three seeds, error bars show sample standard deviation, and points show individual seed results. The +entropy condition is the clearest positive signal; the combined condition has wider observed spread and weaker evidence.}
\label{fig:multiseed_uncertainty}
\end{figure}

Figure~\ref{fig:multiseed_uncertainty} visualizes the same archived SHARDED values as Table~\ref{tab:multiseed}. At 200 steps, +entropy remains +1.0pp over baseline (68.0\% vs.\ 67.0\%), suggesting that the relative gain is not only a 100-step artifact.

\subsection{Training Duration Analysis}

\label{sec:training_duration}

We sweep training steps (100--400) on the 1.7B baseline to validate our choice of 100 steps as the primary operating point.

\begin{table}[t]
\centering
\caption{Step count sweep on Qwen3-1.7B baseline (seed=42). SHARDED peaks at 200 steps, then decreases while the FULL-SHARDED gap widens.}
\label{tab:step_sweep}
\begin{tabular}{lccc}
\toprule
\textbf{Steps} & \textbf{FULL} & \textbf{SHARDED} & \textbf{FULL$-$SH} \\
\midrule
100 & 76.5\% & 66.5\% & 10.0pp \\
200 & 76.5\% & \textbf{68.5\%} & 8.0pp \\
300 & 75.0\% & 64.0\% & 11.0pp \\
400 & 80.5\% & 64.0\% & 16.5pp \\
\bottomrule
\end{tabular}
\end{table}

SHARDED peaks at 200 steps (68.5\%) and then falls, while FULL rises at 400 steps and the FULL-SHARDED gap widens to 16.5pp. We use 100 steps as the primary operating point because it is near the SHARDED peak while halving training time for cross-scale comparisons.

We further validate at 200 steps on the 4B model, shown in Table~\ref{tab:200steps}.

\begin{table}[t]
\centering
\caption{Training stability at 200 steps, 4B, seed=42.}
\label{tab:200steps}
\begin{tabular}{lccc}
\toprule
\textbf{Config} & \textbf{FULL} & \textbf{SHARDED} & $\Delta$SH vs 100-step \\
\midrule
baseline & 79.0\% & 67.0\% & +2.5pp \\
+entropy  & 80.0\% & 68.0\% & +2.0pp \\
combined  & 79.5\% & 67.5\% & --$^\dagger$ \\
\bottomrule
\end{tabular}
\end{table}

$^\dagger$The combined configuration's 100-step reference (71.5\%) was identified as the +7.0pp outlier discussed in \S\ref{sec:outlier}. At 200 steps, combined achieves 67.5\%, consistent with the multi-seed mean of 65.5\% at 100 steps (Table~\ref{tab:multiseed}).

\section{Analysis}

\subsection{Mechanism Scope}

\label{sec:mechanism_scope}

The archived results support a narrow empirical claim: in the LiC/Qwen3 settings evaluated here, entropy masking is the most consistent positive intervention among the tested configurations. They do not establish that entropy masking is generally optimal for OPSD, that $\beta=0.8$ is the best masking ratio, or that overconfidence suppression is the unique cause of the gains.

The results are compatible with three mechanisms:
\begin{enumerate}[leftmargin=*]
    \item \textbf{Overconfidence suppression}. Masking may remove some confidently wrong low-entropy updates induced by dirty history.
    \item \textbf{Signal-to-noise filtering}. Low-entropy positions may be dominated by routine tokens, so filtering them can concentrate GJS loss on more informative disagreements.
    \item \textbf{Implicit curriculum}. High-entropy tokens may mark harder decision points, such as operators, variable bindings, and logical connectors.
\end{enumerate}
These interpretations fit the aggregate pattern, including the smaller gains at larger scales, but they require token-level calibration logs to verify.

Alternative explanations remain plausible: reduced effective supervision, changed gradient scale from the $1/|\mathcal{T}_t|$ normalization, or token-distribution artifacts unrelated to confidence. Distinguishing these accounts requires at least four controls: random masking, divergence-based masking, fixed-denominator masking, and direct entropy-error calibration. We therefore treat the mechanism discussion as interpretation rather than causal evidence.

\subsection{Why Outcome Weighting Is Fragile}

\label{sec:why_outcome_fails}
Outcome correctness scaling assumes that intermediate turn quality correlates with final-answer correctness. In multi-turn OPSD this proxy is noisy: a good intermediate turn can occur in a failed trajectory, while a flawed step can be followed by recovery. Unlike single-turn preference optimization, an error at turn $t$ changes the input distribution at turn $t+1$, so the final outcome is a delayed and entangled signal. A single binary label then applies the same scalar to all middle turns, amplifying variance rather than localizing correction. The EMA teacher already supplies clean-context token targets, so the scalar outcome signal can also conflict with the teacher-student divergence.

This helps explain why the outcome scalar is harmful without masking at 1.7B and inconsistent when combined with masking. Outcome signals may still help at larger scales or with finer granularity, such as turn-level correctness or process feedback, but our results align with credit-assignment work showing that coarse trajectory outcomes are insufficient for OPD without more local signals~\citep{shen2026credit,yang2026ogls}.

\subsection{The Outlier Problem in Small-Scale OPSD}

\label{sec:outlier}

Single-seed OPSD experiments can produce spuriously large effects. The 4B combined outlier (+7.0pp) is far larger than the replicated +1.3pp mean, and the 1.7B combined runs differ by 2.5pp SHARDED despite identical hyperparameters. This matches broader reports that OPD is sensitive to teacher choice, loss formulation, and privileged-information design~\citep{zhu2026manyfaces}. Core OPSD comparisons should therefore include multi-seed results or at least within-configuration replication, and single-seed gains comparable to observed run-to-run variation should remain provisional.

\section{Conclusion}

We presented \textsc{SMOPD}, a selective token-entropy masking variant for dirty-history multi-turn OPSD. In LiC/Qwen3 experiments, \textsc{SMOPD} improves SHARDED accuracy by 1.0--2.5pp across the evaluated 1.7B--8B scales while adding no parameters and zero inference overhead. Outcome-guided correctness scaling is harmful without masking at 1.7B ($-$4.0pp) and inconsistent when combined with masking (+1.3pp at 4B, neutral at 1.7B, $-$0.5pp at 8B).

The practical implication is cautious but useful: start with token-level uncertainty before adding scalar trajectory outcomes. A small step sweep, $\beta=0.8$ as an initial operating point, target-scale validation for any outcome signal, and multi-seed reporting are the most important evaluation practices suggested by these runs.

\subsection*{Practical Takeaways}

\begin{enumerate}[leftmargin=*]
    \item \textbf{Use entropy masking as the first stabilization check}. It is a drop-in loss modification with zero inference cost.
    \item \textbf{Start from $\beta=0.8$, not as a default law}. Treat the ratio as an operating point to sweep when compute allows.
    \item \textbf{Do not rely on binary outcome scaling alone}. The 1.7B degradation is practically meaningful, and the combined effect must be validated at the target scale.
    \item \textbf{Report replication}. The 4B outlier shows that single-seed OPSD gains can be misleading.
\end{enumerate}

\subsection*{Limitations}

The evidence is limited to LiC, Qwen3-Instruct models, math-heavy held-out evaluation, LoRA rank 64, and mostly 100-step training. Cross-scale results are single-seed except for the 4B reliability check, and $\beta=0.8$ is a fixed operating point rather than an established optimum. We also do not report unarchived token-level diagnostics as quantitative evidence; stronger claims about entropy-error calibration require preserved raw token logs and masking controls.

\subsection*{Data, Code, and Compute Availability}

We use the public LiC benchmark and publicly released Qwen3-Instruct base models. This arXiv package includes manuscript source and figures only; model weights, checkpoints, raw logs, and raw evaluation outputs are not included. Quantitative claims are based on archived aggregate artifacts, with fuller code/configuration release planned after cleanup. Experiments used one NVIDIA RTX PRO 6000 Blackwell GPU (96GB VRAM), approximately 40--50 minutes per 100-step run and 60--80 minutes per two-view evaluation.

\subsection*{Broader Impact}

If the result transfers, entropy masking could make multi-turn OPSD slightly more data-efficient without changing inference behavior. We do not identify risks specific to the masking rule beyond the general risks of deploying more capable LLMs.

\medskip

\begin{small}
\noindent\textbf{Acknowledgment}. The authors thank the maintainers of the LiC benchmark for providing standardized multi-turn evaluation protocols.
\end{small}

\FloatBarrier

\clearpage
\section*{Appendix}
\FloatBarrier

\subsection*{A. Reproducibility and Experimental Protocol}
\FloatBarrier

Table~\ref{tab:appendix_protocol} consolidates the experimental settings used for the reported runs. These settings are collected here to make the arXiv version auditable without requiring readers to reconstruct the protocol from the method and results sections.

\begin{table}[!htbp]
\centering
\small
\caption{Protocol and hyperparameter summary for the reported experiments.}
\label{tab:appendix_protocol}
\begin{tabular}{@{}L{0.25\textwidth}L{0.64\textwidth}@{}}
\toprule
\textbf{Component} & \textbf{Setting used in this paper} \\
\midrule
Benchmark & LiC combined benchmark. Training uses 12,165 multi-turn examples; evaluation uses 200 held-out math examples. \\
Evaluation views & FULL is the complete single-turn task view. SHARDED reveals the same task progressively and is the primary multi-turn metric. \\
Models & Qwen3-Instruct at 1.7B, 4B, and 8B scales. \\
Fine-tuning & LoRA rank $r=64$, LoRA $\alpha=128$, dropout $0.0$, applied to attention and feed-forward linear projections. \\
Optimizer & AdamW, learning rate $5\times 10^{-6}$, 100 training steps unless a step sweep or 200-step validation is explicitly reported. \\
Teacher update & EMA teacher initialized from the student and updated with decay $\lambda=0.99$. \\
Middle-turn loss & Generalized Jensen-Shannon divergence with $\beta_{\text{mid}}=0.5$, computed by teacher forcing on the student-generated token sequence. \\
Answer/FULL losses & Reverse KL on the answer turn and on the FULL-preservation branch. The FULL branch is sampled with probability $\rho=0.2$. \\
Pointwise clipping & Per-token GJS and RKL terms are clipped at $\epsilon_{\text{clip}}=0.5$ before aggregation. \\
Entropy masking & For +entropy and combined runs, retain the top 80\% highest-entropy token positions within each eligible middle-turn reply; the first 33\% of training steps disable masking. \\
Outcome weighting & Baseline and +entropy use the reliability proxy in Eq.~\ref{eq:reliability_proxy} with $\eta=0$. For +outcome and combined runs, sensitivity $\eta=0.2$ multiplies this proxy by $1.2$ for incorrect final outcomes and $0.8$ for correct outcomes. \\
Hardware & Single NVIDIA RTX PRO 6000 Blackwell GPU with 96GB VRAM. \\
\bottomrule
\end{tabular}
\end{table}
\FloatBarrier

All reported accuracy values are aggregate evaluation results. The arXiv source package contains the manuscript, style file, and figures needed for compilation, but not model checkpoints, model weights, raw logs, or raw evaluation outputs.

\subsection*{B. Source-to-Claim Provenance}
\FloatBarrier

Table~\ref{tab:appendix_provenance} maps the paper's main quantitative claims to the archived aggregate artifacts used for this arXiv draft. This table is intended as a provenance checklist rather than a replacement for a future full reproducibility release.

\begin{table}[!htbp]
\centering
\small
\caption{Source-to-claim map for reported quantitative evidence.}
\label{tab:appendix_provenance}
\begin{tabular}{@{}L{0.28\textwidth}L{0.33\textwidth}L{0.28\textwidth}@{}}
\toprule
\textbf{Claim or artifact} & \textbf{Reported evidence} & \textbf{Archived source/status} \\
\midrule
1.7B main ablation and Figures~\ref{fig:ablation}, \ref{fig:delta} & Baseline, +entropy, +outcome, and combined aggregate FULL/SHARDED values. & Phase 1 and Phase 2 aggregate artifacts; audited in \url{runs/PAPER_REPORTED_RESULTS_AUDIT_20260729.md}. \\
Cross-scale entropy comparison and Figure~\ref{fig:cross_scale} & Qwen3-1.7B/4B/8B baseline vs.\ +entropy SHARDED deltas. & Phase 1, Phase 2, and Phase 3 aggregate artifacts; figure values listed in \url{my-research-paper/figures/figure_data_manifest.json}. \\
Combined scale summary & 1.7B combined mean, 4B multi-seed combined mean, and 8B single-seed combined result. & Derived reporting table from retained aggregate runs, not a separate experiment. \\
4B multi-seed validation and Figure~\ref{fig:multiseed_uncertainty} & Three-seed baseline, +entropy, and combined SHARDED means with Welch test summaries. & Canonical seed policy and arithmetic checks in \url{runs/DATA_NOTES.md} and \url{runs/PAPER_REPORTED_RESULTS_AUDIT.md}. \\
1.7B step-count sweep and Figure~\ref{fig:step_sweep} & 100, 200, 300, and 400 step FULL/SHARDED results. & Phase 1 aggregate evaluation files and figure manifest. \\
4B 200-step validation & Baseline, +entropy, and combined aggregate FULL/SHARDED results at 200 steps. & Phase 4 aggregate evaluation files; audited in \url{runs/PAPER_REPORTED_RESULTS_AUDIT.md}. \\
Method schematic & Dirty-history student rollouts, clean-context EMA teacher forcing, entropy masking, and unmasked answer/FULL branches. & Conceptual figure generated from the method definition; no additional quantitative claim. \\
\bottomrule
\end{tabular}
\end{table}
\FloatBarrier

The current manuscript deliberately excludes unarchived token-level diagnostics, per-turn slices, prompt-level examples, qualitative case studies, and masking-ratio sensitivity rows as quantitative evidence. Those analyses may be useful in a later release, but they should be reported only after the corresponding raw artifacts are preserved.

\subsection*{C. Statistical Interpretation Notes}
\FloatBarrier

The held-out evaluation set contains 200 examples, so reported accuracies move in increments of 0.5 percentage points. A 1.0pp SHARDED difference corresponds to two evaluation examples, and a 2.5pp difference corresponds to five examples. For this reason, the single-seed cross-scale comparisons in Table~\ref{tab:cross_scale} should be read as directional descriptive evidence, not as precise effect-size estimates.

The 4B reliability check uses three seeds per configuration. Table~\ref{tab:multiseed} reports Welch's two-tailed $t$-test for SHARDED accuracy against the baseline; the corresponding directional one-tailed values are discussed in the main text because the intervention hypothesis is directional. With $n=3$, these $p$-values are rough indicators under limited power. They support a cautious reliability claim for +entropy in this setting, but they do not establish a general scaling law or an optimal masking ratio.

The combined configuration is handled conservatively because an early 4B seed-42 run produced a +7.0pp SHARDED outlier. That run is reported in the outlier discussion and is not folded into the multi-seed reliability mean. The combined row in Table~\ref{tab:multiseed} therefore estimates the subsequent reliability runs, while the outlier remains visible as a warning about single-seed OPSD variance.

\begin{table}[!htbp]
\centering
\small
\caption{How the main empirical results should be interpreted.}
\label{tab:appendix_interpretation}
\begin{tabular}{@{}L{0.27\textwidth}L{0.29\textwidth}L{0.34\textwidth}@{}}
\toprule
\textbf{Evidence item} & \textbf{Supported reading} & \textbf{Boundary} \\
\midrule
1.7B ablation & Entropy masking is the strongest positive intervention in the primary ablation. & Single seed for baseline/+entropy/+outcome; combined is averaged over two archived runs. \\
Cross-scale +entropy & Positive SHARDED deltas appear at 1.7B, 4B, and 8B. & Descriptive single-seed evidence; scale trend should not be read as a law. \\
4B multi-seed check & +entropy has the clearest replicated signal among tested variants. & Only three seeds; $p$-values are rough indicators under limited power. \\
Combined outcome weighting & The outcome scalar may contain useful signal at 4B but is inconsistent overall. & 1.7B and 8B results do not support a general combined benefit. \\
Step sweep & SHARDED is non-monotonic over training steps. & The sweep is on 1.7B baseline only; it motivates calibration rather than an optimal schedule. \\
\bottomrule
\end{tabular}
\end{table}
\FloatBarrier

\subsection*{D. Implementation Notes Relative to Prior Work}
\FloatBarrier

The method is closest to the MAIGO/REOPOLD line of work, but it is not an exact reproduction of either algorithm. Relative to MAIGO, we use a clean-teacher / dirty-student training family, but average over all eligible non-empty middle turns in a rollout rather than sampling one eligible turn as an estimator. The detached reliability proxy in Eq.~\ref{eq:reliability_proxy} is common to the reported configurations; the outcome variant multiplies this proxy by a correctness scalar and is not MAIGO's full adaptive reliability estimator.

Relative to REOPOLD, the entropy selection rule is adapted to multi-turn OPSD rather than single-turn OPD refinement. The threshold is computed within each generated middle-turn reply, not as a batch-level percentile. The REOPOLD-inspired schedule in Algorithm~\ref{alg:training} only delays entropy masking during the first 33\% of training steps; it does not implement REOPOLD's reward-based filtering phase.

All divergence terms are computed with token alignment through teacher forcing: the student and EMA teacher are evaluated on the same student-generated token sequence, while their contexts differ. This avoids aligning two independently generated responses. Entropy masking applies only to middle-turn GJS terms; answer-turn RKL and FULL-preservation RKL are left unmasked.

\subsection*{E. Per-Seed Breakdown for Multi-Seed Validation}
\FloatBarrier

Table~\ref{tab:appendix_perseed} lists the exact rows used for the 4B reliability summary. It is a provenance breakdown rather than a strictly paired-seed design: baseline and +entropy use the paper-canonical Phase 3 seed-42 rows, while the combined seed-42 row is a subsequent reliability rerun and excludes the earlier +7.0pp outlier.

\begin{table}[!htbp]
\centering
\caption{Per-seed results for multi-seed validation (4B, 100 steps).}
\label{tab:appendix_perseed}
\begin{tabular}{lcccc}
\toprule
\textbf{Config} & \textbf{Seed} & \textbf{FULL} & \textbf{SHARDED} \\
\midrule
\multirow{3}{*}{baseline} & 42   & 78.5\% & 64.5\% \\
                          & 123  & 77.5\% & 64.5\% \\
                          & 456  & 78.5\% & 63.5\% \\
\midrule
\multirow{3}{*}{+entropy} & 42   & 79.5\% & 66.0\% \\
                          & 123  & 80.0\% & 65.5\% \\
                          & 456  & 79.5\% & 66.0\% \\
\midrule
\multirow{3}{*}{combined$^{\,\ddagger}$} & 42   & 79.0\% & 65.0\% \\
                          & 123  & 77.5\% & 66.5\% \\
                          & 456  & 79.0\% & 65.0\% \\
\bottomrule
\end{tabular}
\par\vspace{2pt}
\begin{minipage}{0.72\textwidth}
\footnotesize
$^\ddagger$For baseline and +entropy, seed 42 is the cross-scale run used in Table~\ref{tab:cross_scale}. The combined seed-42 row is a subsequent reliability rerun; the earlier combined seed-42 outlier is reported separately in \S\ref{sec:outlier}.
\end{minipage}
\end{table}
\FloatBarrier

\subsection*{F. Step Sweep Figure}
\FloatBarrier

\begin{figure}[!htbp]
\centering
\includegraphics[width=0.78\textwidth]{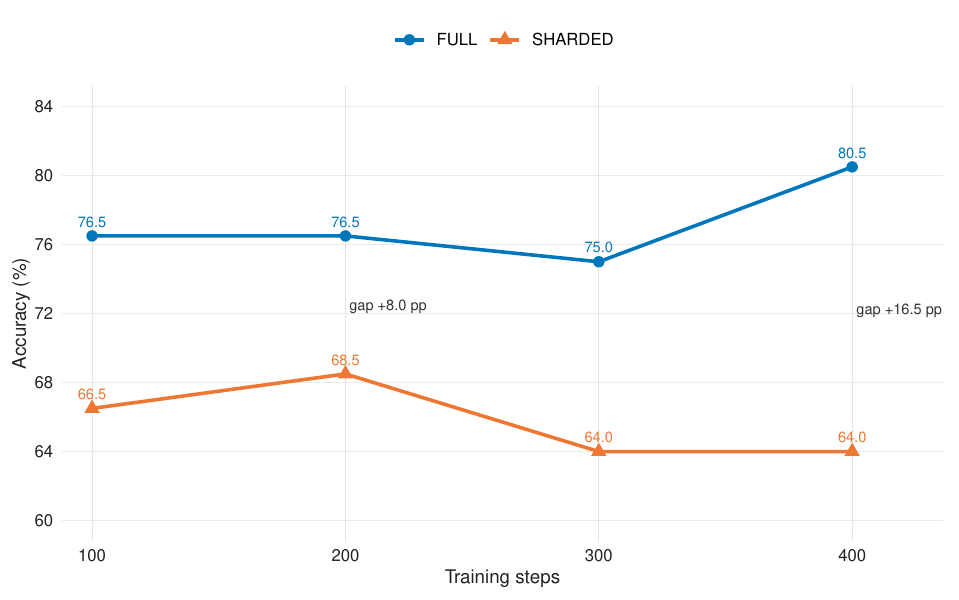}
\caption{\textbf{Step count sweep on 1.7B baseline.} SHARDED peaks at 200 steps then decreases, while FULL rises at 400 steps, suggesting possible overfitting to single-turn performance.}
\label{fig:step_sweep}
\end{figure}

\FloatBarrier

\end{document}